\documentclass[10pt,twocolumn,letterpaper]{article}

\usepackage{iccv}
\usepackage{times}
\usepackage{epsfig}
\usepackage{graphicx}
\usepackage{amsmath}
\usepackage{amssymb}
\usepackage{balance}
\usepackage{gensymb}
\usepackage[table,xcdraw]{xcolor}
\usepackage[colorlinks,breaklinks=true,bookmarks=false]{hyperref}

\iccvfinalcopy 

\def\iccvPaperID{6215} 

\begin{document}

\title{Self-Supervised Deep Depth Denoising}

\author{Vladimiros Sterzentsenko \thanks{Equal contribution}
\quad \quad \quad
Leonidas Saroglou \footnotemark[1]
\quad \quad \quad
Anargyros Chatzitofis \footnotemark[1]\\
\quad
Spyridon Thermos \footnotemark[1]
\quad \quad \quad
Nikolaos Zioulis \footnotemark[1]
\quad \quad \quad
Alexandros Doumanoglou\\
\quad
Dimitrios Zarpalas 
\quad \quad \quad
Petros Daras
\\
\fontsize{10}{12}\selectfont Information Technologies Institute (ITI), Centre for Research and Technology Hellas (CERTH), Greece\\
}

\maketitle

\begin{abstract}
Depth perception is considered an invaluable source of information for various vision tasks.
However, depth maps acquired using consumer-level sensors still suffer from non-negligible noise.
This fact has recently motivated researchers to exploit traditional filters, as well as the deep learning paradigm, in order to suppress the aforementioned non-uniform noise, while preserving geometric details.
Despite the effort, deep depth denoising is still an open challenge mainly due to the lack of clean data that could be used as ground truth.
In this paper, we propose a fully convolutional deep autoencoder that learns to denoise depth maps, surpassing the lack of ground truth data. 
Specifically, the proposed autoencoder exploits multiple views of the same scene from different points of view in order to learn to suppress noise in a self-supervised end-to-end manner using depth and color information during training, yet only depth during inference. 
To enforce self-supervision, we leverage a differentiable rendering technique to exploit photometric supervision, which is further regularized using geometric and surface priors. 
As the proposed approach relies on raw data acquisition, a large RGB-D corpus is collected using Intel RealSense sensors. 
Complementary to a quantitative evaluation, we demonstrate the effectiveness of the proposed self-supervised denoising approach on established 3D reconstruction applications. Code is avalable at \url{https://github.com/VCL3D/DeepDepthDenoising}
\end{abstract}

\def\pixel{\mathbf{p}}
\def\R{\mathbb{R}}
\def\vertex{\mathbf{v}}
\def\normal{\mathbf{n}}
\def\depthmap{D}
\def\normalsmap{\mathbf{N}}
\def\colormap{\mathbf{I}}
\def\T{\mathbf{T}}
\def\K{\mathbf{K}}
\def\loss{\mathcal{L}}
\newcommand{\fromto}[2]{{#1} \rightarrow {#2}}
\definecolor{limegreen}{rgb}{0.3, 0.85, 0.3}

\section{Introduction}
\begin{figure}[h]
    \centering
    \includegraphics[width=\columnwidth]{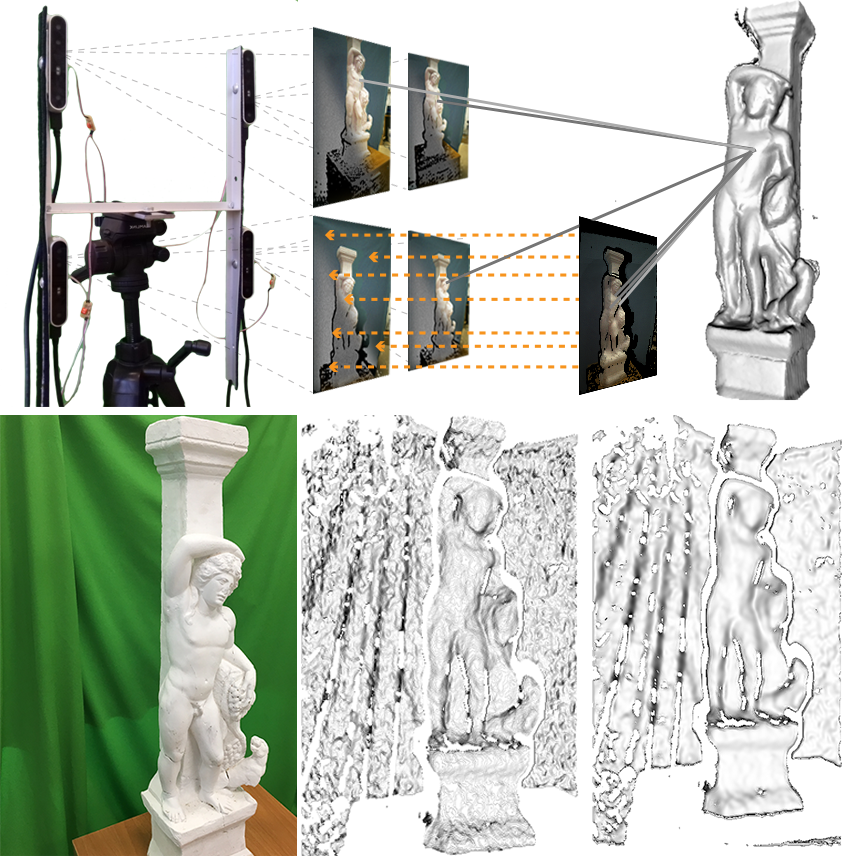}
    \caption{An abstract representation of the proposed method. Our model exploits depth-image-based rendering in a multi-view setting to achieve self-supervision using photometric consistency and geometrical and surface priors. A denoising example is visualized in the lower part (right-most), compared to a traditional filtering result (middle).}
    \label{fig:intro_concept}
    \vspace{-0.2in}
\end{figure}

Depth sensing serves as an important information cue for all vision related tasks. 
Upon the advent of consumer grade depth sensors, the research community has exploited the availability of depth information to make performance leaps in a variety of domains. 
These include SLAM technology for robotics navigation,  static scene capture or tracking for augmented reality applications \cite{tateno}, dynamic human performance capture \cite{alexiadis}, autonomous driving \cite{chen}.

Depth sensors can be categorized based on either their interaction with the observed scene in \textit{passive} (pure observation) and \textit{active} (observation after actuation), or their technological basis in \textit{stereo}, \textit{structured light} (SL) and \textit{time-of-flight} (ToF) respectively.
While the latter two are active by definition, stereo-based sensors can operate in both passive and active mode as they estimate depth via binocular observation and triangulation.
Given that they are driven by correspondence establishment, the active projection of textured patterns into the scene improves performance in low textured areas.
However, the aforementioned sensor types suffer from high levels of noise and structural artifacts.\\
\indent Most works that aim to address noisy depth estimations rely on using traditional filtering methods \cite{Matsumoto2015, Zhang2014}, explicit noise modeling \cite{shen2013layer, herrera2012joint, basso2018robust}, and the exploitation of the Deep Learning (DL) paradigm in terms of deep denoising autoencoders. 
However, the former two require extensive parameter tuning to properly adapt to different levels of noise, struggle to preserve details, and lead to local (sensor-specific) solutions. 
On the other hand, recent studies utilizing deep autoencoders \cite{Jeon2018, Yan2018} are able to capture context and lead to more global solutions. 
The main challenge with the data-driven approaches is that finding ground truth for supervision is a hard, time-consuming, usually expensive process and sometimes impossible. 
Although more recent unsupervised data-driven approaches \cite{Naumova2017} try to address the ground truth drawback, they rely on assumptions for the noise nature and properties, which do not apply to consumer level depth sensors.\\
\indent In this work, the DL paradigm is adopted to address both the lack of ground truth data, as well as the necessity to investigate denoising without a priori assumptions.
A fully-convolutional deep autoencoder is designed and trained following a self-supervised approach. 
In particular, self-supervision relies on simultaneously capturing the observed scene from different viewpoints, using multiple RGB-D sensors placed in a way that their fields of view (FoV) overlap. 
The color information acquired by the sensors is used for synthesizing target view using the predicted depth maps given known sensor poses.
This process enables direct photometric supervision without the need for ground truth depth data. 
Depth and normal smoothness priors are used for regularization during training, while our inference only requires a single depth map as input.
The model is trained and evaluated on a corpus collected with the newest Intel RealSense sensors \cite{keselman2017intel} and consists of sparse data with high depth variation. However, note that on inference, the model can be applied to any consumer-level depth sensor. An overview of our method, along with a denoising example are depicted in Fig.~\ref{fig:intro_concept}.

Extensive quantitative evaluation demonstrates the effectiveness of the proposed self-supervised denoising method compared to state-of-the-art methods. Additionally, the performance of the deep autoencoder is further evaluated qualitatively by using the denoised depth maps in well-established 3D reconstruction applications, showcasing promising results given the noise levels of the utilized sensors. 
Note that the model structure enables efficient inference on recent graphics cards.

\section{Related Work}
Each depth sensing technology is affected with distinct systematic noise, a fact that renders the development of universal depth denoising methods a challenging task. 
In the following overview, related work is divided in three major categories, presenting state-of-the-art depth denoising approaches available in the literature.
\par
\textbf{Noise modeling.} 
As depth sensors operate on different principles, they are also affected by different systematic noise that is unique to their underlying operation.
As a result, one approach of addressing the levels of noise in depth maps is to model the underlying sensor noise.
The initial work of \cite{herrera2012joint} modeled Kinect's systematic noise into a scale and a distortion component, and was solved as a combined problem of noise modeling, extrinsic and intrinsic calibration, using planar surfaces and a checkerboard.
In addition to denoising, \cite{shen2013layer} also performed depth map completion on data produced by a SL depth sensor. A probabilistic framework for foreground-background segmentation was employed, followed by a neighbourhood model for denoising which prevented depth blurring along discontinuities. 
A similar approach was recently proposed by \cite{basso2018robust}, with the key difference being a polynomial undistortion function which was estimated in a finer granularity at the pixel level rather than a closed form equation.
However, the heterogeneity of sensor noise models is difficult to generalize and apply in a variety of sensors.
A prominent example is a bulk of recent work that deals with the noise inducing multiple path interference (MPI) issue of ToF sensors \cite{guo2018tackling}, \cite{Marco2017} and \cite{Agresti2019}. 
They employ DL methods to correct and denoise the generated depth data, but these approaches are not applicable to other sensor types.

\textbf{Classical and Guided Filtering.}
Traditional filtering approaches are more applicable to a variety of sensor types, with the most typical approach for depth denoising in various applications (\textit{e.g.}~ \cite{newcombe2011kinectfusion}) being the
bilateral filter \cite{Tomasi}, a well established computer vision filter.
From a more practical standpoint, as depth sensors are typically accompanied by at least one light intensity sensor (color, infrared), many works have resorted to using this extra modality as a cleaner guidance signal to drive the depth denoising task. 
While indeed a promising approach, the use of intensity information relies on the aligned edge assumption between the two modalities, and as a result, both the joint bilateral \cite{Matsumoto2015} and rolling guidance \cite{Zhang2014} filters suffer from texture transfer artifacts.
Thus, follow up works have focused on addressing this lack of structural correlation between the guide and the target images \cite{Shen2015, Ham2015, Lu2014}.
Finally, similar in concept approaches \cite{OrEl2015, Han2013, Wu2014, Yu2013} utilize shading information, extracted from the intensity images, in order to refine the acquired depth maps. 
Despite the increased robustness gained from surface information utilization, all aforementioned methods cannot alleviate from artifacts produced due to modalities misalignment. 
Additionally, the most significant drawback of typical filtering is its inability to understand the global context, thus operating on local level. 

\textbf{Learning methods.}
Data driven methods on the other hand, can better capture the global context of each scene, an important source of information that can drive the denoising task.
The guided filtering concept has been implemented with convolutional neural networks (CNNs) in \cite{Gu2017} and \cite{Li2019}. 
The former proposes a weighted analysis representation model in order to model the dependency between intensity and depth images, with a local weight function learned over labeled task-specific data.
The latter currently represents the state-of-the-art in joint filtering. 
It uses 3 CNNs to learn to transfer structural information from the guiding image to the noisy one.
While effective, it is learned in a fully supervised manner, meaning that it requires ground truth data, which are hard to obtain and would require collection for each different type of sensor. 
More recent works have resorted to near ground truth dataset generation in order to circumvent the lack of and difficulty in acquiring ground truth depth data.
The ScanNet \cite{Dai2017} dataset is used in \cite{Jeon2018} to produce raw-clean depth pairs by exploiting the implicitly denoised 3D reconstructed models and the known sensor poses during scanning, to synthesize them via rendering.
A very deep multi-scale Laplacian pyramid based auto-encoder model is used and directly supervised with an additional gradient-based structure preserving loss.
Despite the satisfactory results, inference is quite slow because of the depth of their network, making it unfeasible to use their model in real-world applications.
Similarly, Kwon \etal \cite{HyeokHyenKwon2015} produce their raw-near ground truth pairs using \cite{newcombe2011kinectfusion}, in order to train their multi-scale dictionary-based method. 
Additonally, Wu \etal \cite{Yan2018} use a dynamic 3D reconstruction method \cite{guo2017real} to non-rigidly fuse depth data and construct raw-clean depth map pairs. 
This work employs an auto-encoder with skip connections, coupled with a refinement network at the end that fuses the denoised data with intensity information to produce refined depth maps.

Unavailability and difficulty to generate ground-truth data in various contexts, is the major motivator for unsupervised methods. 
Noise2Noise \cite{Naumova2017} and its extensions Noise2Self \cite{DBLP:journals/corr/abs-1901-11365} and Noise2Void \cite{DBLP:journals/corr/abs-1811-10980} demonstrated how denoising can be achieved in an unsupervised manner without clean data. 
However the aforementioned approaches rely on certain distributional assumptions (\textit{i.e.}~ zero-mean Gaussian i.i.d. noise), which do not apply on data acquired by consumer-level depth sensors. \color{black}
Evidently, methods for training without direct supervision are required. Our work addresses this issue by proposing the use of multiple sensors in a multi-view setting.

\section{Depth Denoising}\label{sec::approach}
\begin{figure}[t]
    \centering
    \includegraphics[width=0.9\columnwidth]{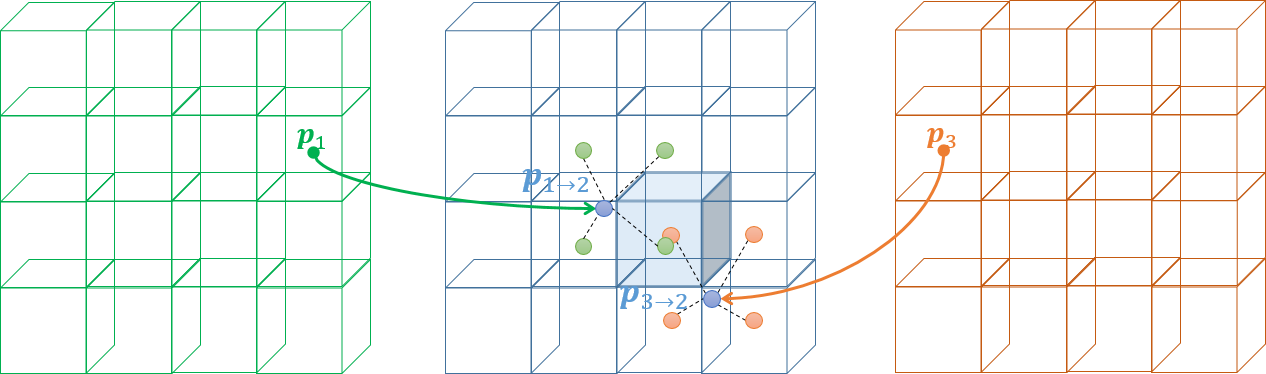}
    \caption{Our multi-view forward splatting scheme is illustrated. The source views - 1 and 3 (\textit{\textcolor{limegreen}{green}} and \textit{\textcolor{orange}{orange}} respectively) - splat their contributions to the target view - 2 (\textit{\textcolor{blue}{blue}}). Each source pixel ($\textcolor{limegreen}{\pixel_1}$ and $\textcolor{orange}{\pixel_3}$) reprojects to the target view ($\textcolor{blue}{\pixel_{\fromto{1}{2}}}$ and $\textcolor{blue}{\pixel_{\fromto{3}{2}}}$ respectively). The color information that they carry from their source views is spread over the neighborhood of their reprojections in a bilinear manner. In addition, these contributions are also weighted by each source pixel's confidence. As shown in the highlighted pixel of target view, multiple views combine their color information in the target splatted image.
    }
    \label{fig:splatting}
    \vspace{-0.1in}
\end{figure}

\begin{figure*}[t]
    \centering
    \includegraphics[width=.85\textwidth]{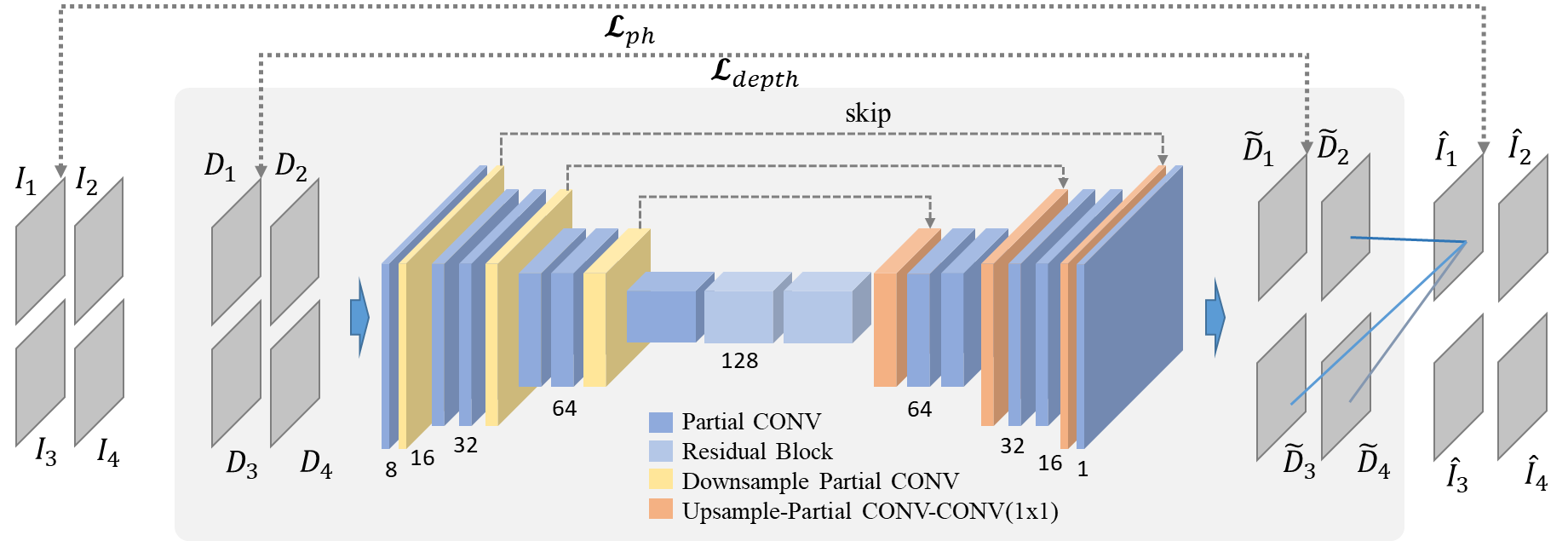}
    \caption{Detailed network architecture of the proposed depth denoising method. The network receives raw depth information from all available sensors (${D}_{1}-{D}_{4}$) and predicts denoised depthmaps ($\tilde{D}_{1}-\tilde{D}_{4}$). Using differentiable rendering (see Section \ref{supervision}), a new target color image $\hat{\colormap}_{1}$ is synthesized from the non-target depth map predictions ${D}_{2}-{D}_{4}$. Subsequently, $\hat{\colormap}_{1}$ is used to compute the $L_{ph}$ loss (see Section \ref{losses}), considering $\colormap_{1}$ as ground truth. Note that every input depth map is iteratively considered as target frame, while the total loss derives from the summation of each sensor loss.}
    \label{fig:net}
    \vspace{-0.1in}
\end{figure*}

Our approach aims to circumvent the lack of ground truth depth data. 
An end-to-end framework, trained in the absence of clean depth measurements, learns to denoise the input depth maps.
Using unstructured multi-view sensors that capture unlabelled color and depth data, our approach relies on view synthesis as a supervisory signal and, although utilizing color information during training, it requires only a single depth map as input during inference.

\subsection{Multi-view Self-Supevision}
\label{supervision}

Each sensor jointly acquires a color image $\colormap(\pixel)\in\R^3$ and a depth map $D(\pixel)\in\R$, with $\pixel:= (x,y) \in \Omega$ being the pixel coordinates in the image domain $\Omega$ defined in a $W \times H$ grid, with $W$ and $H$ being its width and height, respectively.
Considering $\mathcal{V}$ spatially aligned sensors $v\in\{1,...,\mathcal{V}\}$, whose viewpoint positions are known in a common coordinate system  and expressed by their poses $\T_v:= \left[ \begin{smallmatrix} \mathbf{R}_v&\mathbf{t}_v\\ \mathbf{0}&1 \end{smallmatrix} \right]$, where $\mathbf{R}_v$ and $\mathbf{t}_v$ denote rotation and translation respectively, we can associate image domain coordinates from one viewpoint to another using: 
\begin{equation}
\label{eq:mapping}
    \mathcal{T}_{\fromto{s}{t}}(\pixel_s) =  \pi(\T_{\fromto{s}{t}}\pi^{-1}(D_s(\pixel_s), \K_s), \K_t),
\end{equation}
with $\T_{\fromto{s}{t}}$ being the relative pose between sensors $s$ (source) and $t$ (target), with the arrow showing the direction of the transformation. $\pi$ and $\pi^{-1}$ are the projection and deprojection functions that transform 3D coordinates to pixel coordinates and vice versa, using each sensor's intrinsics matrix $\mathbf{\K}$. Note that we omit the depth map $D_s$, pose $\T_{\fromto{s}{t}}$ and the intrinsics $\K_s$ and $\K_t$ arguments from function $\mathcal{T}$ for notational brevity.

Under a multi-view context and given that each $v$ sensor color image $\colormap_v$ and depth map $D_v$ are aligned and defined on the same image domain $\Omega$, color information can be transferred from each view to the other ones using Eq.~\ref{eq:mapping} on a per pixel basis.
Note that contrary to noisy depth measurements, the acquired color information can be considered as clean (or otherwise a much more consistent and higher quality signal). 
Traversing from one view to another via noisy depth will produce distorted color images due to depth errors manifesting into incorrect reprojections.
Consequently, we can self-supervise depth noise through inter-view color reconstruction under the photoconsistency assumption.
 
Even though view synthesis supervision requires at least 2 sensors, more of them can be employed, as long as their poses in a common coordinate system are known, via the geometric correspondence function $\mathcal{T}$.
This allows us to address apparent issues like occlusions and the limitations of a consistent baseline (restricted accuracy).
Additionally, as the noise is inconsistent, multiple depth maps observing the same scene will simultaneously offer varying noise structure and levels, while increasing the diversity of the data.
Intuitively, and similar to wide-baseline stereo, adding more sensors will offer higher reconstruction accuracy due to the variety of baselines. Note that since this approach is purely geometric, any number of unstructured sensor placements is supported. 

Most works using view synthesis as a supervision signal utilize inverse warping \cite{NIPS2015_5854} for image reconstruction.
Under this reconstruction scheme, each target pixel samples from the source image, thus many target pixels may sample from the same source pixel. 
However, relying on erroneous depth values is problematic as occlusions and visibility need to be handled in an explicit manner via depth testing, itself relying on the noisy depth maps.

To overcome this, we employ differentiable rendering \cite{tulsiani2018layer} and use forward splatting to accumulate color information to the target view.
In forward splatting each source pixel accumulates its contribution to the target image, thus, as depicted in Fig.~\ref{fig:splatting}, many source pixels (from the same or even different views) may contribute to a single target pixel.
This necessitates a weighted average accumulation scheme for reconstructing the rendered image. 
We define a splatting function $\mathcal{S}_{\fromto{s}{t}}(A_t, B_s, D_s, \pixel_s)$:
\begin{equation}
\label{eq:splatting}
    A_t(\mathcal{T}_{\fromto{s}{t}}(\pixel_s)) = w_{c}(D_s, \pixel_s) w_{b}(\mathcal{T}_{\fromto{s}{t}}(\pixel_s), \ddddot{\pixel_t}) B_s(\pixel_s)
\end{equation}
with $A, B$ images defined in $\Omega$, $w_c$ weighting the source pixel's contribution, and $w_b$ being a bilinear interpolation weight as the re-projections of Eq.~\ref{eq:mapping} produce results at sub-pixel accuracy.
Therefore, we ``\textit{splat}" each source pixel to contribute to the re-projected target pixel's immediate neighborhood, expressed by $\ddddot{\pixel_t} \in \{ \, {}_x\lfloor \pixel_t \rfloor_y, {}_x\lfloor \pixel_t \rceil_y, {}_x\lceil \pixel_t \rceil_y, {}_x\lceil \pixel_t \rfloor_y \, \}$, where $\lceil.\rceil$ and $\lfloor. \rfloor$ denote ceiling and floor operations respectively in the sub-scripted image domain directions $x,y$. 
Effectively, every source pixel will splat its contribution to four target pixels, enforcing local differentiability.

We weight the contribution of each pixel by taking its uncertainty into account which is expressed as the combination of the measurement noise along the ray, as well as the radial distortion error: $w_c(D, \pixel) = w_d(D, \pixel) w_r(\pixel)$.
To increase applicability, both the depth uncertainty and the radial distortion confidence weights are modelled in a generic way.
For depth uncertainty, we consider measurements closer to the sensor's origin as more confident than farther ones; $w_d(D, \pixel) = \exp (\frac{-D(\pixel)}{\sigma_D})$, controlled by $\sigma_D$.
Similarly, for the radial distortion confidence a generic FoV model is used \cite{tang2017precision}:
\begin{equation}
    \mathcal{R}(\pixel) = \frac{tan(r(\pixel)tan(\omega))}{tan(\omega)},
\end{equation}
where $r(\pixel) = \sqrt{(x^2+y^2)}$ is the pixel's radius from the distortion center (\textit{i.e.}~the principal point) and $\omega$ is half the sensor's FoV.
In this way, measurements in high distortion areas are considered as less confident and weighted by $w_r(\pixel) = \exp(\frac{r(\pixel)}{\mathcal{R}(\pixel)})$.

We splat and accumulate the weighted color contributions from a source image $s$ to a target image $t$, as well as the weights themselves via the splatting function $\mathcal{S}$:
\begin{equation}
\label{eg:weight_splatting}
    \mathcal{S}_{\fromto{s}{t}}(\hat{\colormap}_t, \colormap_s, D_s, \pixel_s), \qquad
    \mathcal{S}_{\fromto{s}{t}}(W_t, \mathbf{1}, D_s, \pixel_s),
\end{equation}
where $W$ and $\mathbf{1}$ are scalar maps of splatted weights and ones respectively, defined in the image domain $\Omega$.
In order to compute the reconstructed image, a weighted average normalization is performed in the target view; $\hat{\colormap}_t = \hat{\colormap}_t \oslash (W_t \oplus \epsilon)$, with $\epsilon$ being a small constant to ensure numeric stability, while circles denote element wise operators.

Note that through forward splatting, the blended values of the target image enable gradient flow to all contributing measurements. In traditional rendering (discrete rasterization), gradient flow to depth values close to surface would be cut-off, and given the bidirectional nature of noise, this would encumber the learning process. On the contrary, using forward splatting, background depths only minimally contribute to the blended pixels due to the exponential weight factor, receiving minimal gradients, thus implicitly handling occlusions and visibility tests.

In a multi-view setting with $S$ sensors, we can splat the contributions in a many-to-one scheme, in order to fully exploit multi-view predictions.
For each view $t$, a splatted image is rendered by accumulating the color and weight splats from all other views to the zero-initialized $\hat{\colormap}_t, W_t$:
\begin{equation}
\label{eq:multi-view_splatting}
    \forall \{s,t|t\!\neq\!s\}\!\in\!S: \mathcal{S}_{\fromto{s}{t}}(\colormap_s, \hat{\colormap}_t, D_s, \pixel_s),  \mathcal{S}_{\fromto{s}{t}}(W_s, \mathbf{1}, D_s, \pixel_s) 
\end{equation}
and then subsequently $\hat{\colormap}_t$ is normalized.
The presented depth-image-based differentiable rendering allows us to exploit photometric supervision in a many-to-many scheme, thus relying only on aligned color information to supervise depth denoising.

\subsection{Network Architecture}
\label{network}
The proposed data-driven approach is realized as a deep autoencoder depicted in Fig.~\ref{fig:net}. 
Its structure is inspired by the U-Net~\cite{ronn} architecture that consists of an encoder, a latent, and a decoder part, respectively.
Note that the network is fully convolutional as there is no linear layer.

The encoder follows the typical structure of a CNN and consists of 9 convolutional (CONV) layers each followed by an Exponential Linear Unit (ELU) \cite{clevert} activation function. 
The input is downsampled 3 times prior to the latent space using convolution with $3\times 3$ kernels and stride 2, while the number of channels is doubled after every downsampling layer.

The latent part consists of 2 consecutive residual blocks each following the ELU-CONV-ELU-CONV structure adopting the pre-activation technique and the identity mapping introduced in \cite{he} for performance improvement.

The decoder shares similar structure with the encoder, consisting of 9 CONV layers each followed by an ELU non-linearity. 
The features are upsampled 3 times prior to the final prediction, using nearest neighbor upsampling followed by a CONV layer.
Note that each downsampling layer is connected with the corresponding upsampling one (features with the same dimensions) with a skip connection.
Subsequently, the activations of the upsampling layer are concatenated with the ones from the corresponding skip connection. 
After concatenation, a CONV layer with $1\times 1$ kernel size follows, forcing intra-channel correlations learning.

In order to ensure that denoising is not affected either from invalid values due to data sparsity or depth difference in edge-cases, the recently presented partial convolutions \cite{liu} are used in every CONV layer.
The required validity (binary) mask $M$ is formed by parsing the input depth map $D$ and setting $M(\pixel) = 1$ for $D(\pixel) > 0$ and $M(\pixel) = 0$ for zero depth. 
This mask is given as input to the network and is updated after each partial convolution as in \cite{liu}.

During training, the network infers a denoised depth map for each sensor. 
Considering input from 4 sensors, as in Fig.~\ref{fig:net}, all depth maps are iteratively set as target frames. 
Thus, following the forward splatting technique presented in Section \ref{supervision}, target $\hat{\colormap}$ is synthesized using information from the non-target predicted depth maps. 
The target $\colormap$ and $\hat{\colormap}$ are used to compute the photometric loss, which is discussed in the next section. 
Note that the gradients are accumulated for all different target depth maps and the weights update of the network is performed once.
This way we perform denser back-propagation in each iteration, even though our inputs are sparse, leading to faster and smoother convergence.

\subsection{Losses}
\label{losses}
The proposed network is trained using a geometrically-derived photometric consistency loss function. Additionally, depth and normal priors are exploited as further regularization, which force spatial consistency and surface smoothness. The total loss that is used to compute the network gradients is defined as:
\begin{equation}
    \loss_{total} = \underbrace{\lambda_{1}  \loss_{ph}}_\textrm{data} + \underbrace{\lambda_{2} \loss_{depth} + \lambda_{3} \loss_{surface}}_\textrm{priors},
\end{equation}
where $\lambda_{1},\lambda_{2},\lambda_{3}  \in (0,1)$ are hyperparameters that add up to 1. The photometric loss, as well as the regularization functions are discussed in detail below.

\textbf{Photometric consistency}: $\loss_{ph}$ forces the network to minimize the pixel-wise error between input $\colormap$ and $\hat{\colormap}$. Note that in order to perform correct pixel-wise supervision, we compute the binary mask of $\hat{\colormap}$, denoted as $M_{splat}$, where $M_{splat}(\pixel)=1$ for $\hat{\colormap}(\pixel) > 0$ and $M_{splat}(\pixel)=0$ for zero $\hat{\colormap}(\pixel)$ values. Subsequently, the masked input image $\ddot{\colormap}$ is used as ground truth and is computed as $\ddot{\colormap} = M_{splat}\odot{\colormap}$, where $\odot$ denotes element-wise multiplication. $\loss_{ph}$ is composed of two terms, namely the ``color-based" $\loss_{col}$ and the ``structural" $\loss_{str}$ loss, respectively. 
The color-based loss is defined as:
\begin{equation}
\label{eq:color}
\loss_{col} = \sum_{\pixel} \rho(M(\pixel)||\ddot{\colormap}(\pixel) - \hat{\colormap}(\pixel)||_1) ,
\end{equation}
where $M$ is the validity mask (see Section \ref{network}) and $\rho(x) = \sqrt{x^2 + \gamma^2}$ is the Charbonnier penalty \cite{charbo2, charbo} ($\gamma$ is a near-zero constant) used for robustness against outliers. $\loss_{col}$ aims to penalize deviations in the color intensity between $\ddot{\colormap}$ and $\hat{\colormap}$. On the other hand, we use structured similarity metric (SSIM) as the structural loss between $\ddot{\colormap}$ and $\hat{\colormap}$ which is defined as:
\begin{equation}
\loss_{str} = 0.5\sum_{\pixel} \phi(M(\pixel)(1 - \textrm{SSIM}(\ddot{\colormap}(\pixel), \hat{\colormap}(\pixel)))),
\end{equation}
where $M$ is the same validity mask as in Eq.~\ref{eq:color} and $\phi(x)$ is the Tukey's penalty, used as in \cite{belagiannis} given its property to reduce the magnitude of the outliers' gradients close to zero.
Intuitively, $\loss_{str}$ forces prediction invariance to local illumination changes and structural information preservation.
Note that the aforementioned penalty functions are used to address the lack of constrains (\textit{i.e.}~Lambertian surfaces, no occlusions) that need to be met for photometric consistency supervision, albeit not applicable on real-world multi-view scenarios.
Finally, the total photometric loss function is defined as the linear combination of the aforementioned color-based and structural losses, and is given by:
\begin{equation}
    \loss_{ph} = (1-\alpha) \loss_{col} + \alpha \loss_{str},
\end{equation}
where $\alpha \in (0,1)$ is a hyperparameter.

\textbf{Depth regularization.} We choose to further regularize the aforementioned photometric consistency loss by exploiting depth information priors. In particular, considering the residual $r = M \odot (D - \tilde{D})$, where $\tilde{D}$ is the denoised prediction of the network, we use the inverse Huber (BerHu) penalty \cite{berhu}:
\begin{equation}
\loss_{depth} = \left\{
\begin{array}{ll}
     |r|, & |r|\leq c  \\
     \frac{r^2 + c^2 }{2c}, & |r|>c 
\end{array} ,
\right.
\end{equation}
where $c$ is a border value defined as the $20\%$ of the maximum per batch residual $c=0.2\max(r)$. The choice of BerHu instead of  $L2$ is based on \cite{laina2016deeper}, where it was found that it is more appropriate as a depth estimator, as it behaves as $L1$ for residuals lower than the border value.

\textbf{Surface regularization.} Besides depth regularization, a surface regularization prior is used to enforce smoothness in the predicted depth maps. In particular, the surface loss is given by: 
\begin{equation}
\loss_{surface} = 1 - \sum_{\pixel} \\
\sum_{\pixel' \in \Theta_{\pixel} } |\langle \normal(\pixel) , \normal(\pixel') \rangle| \frac{M(\pixel)}{G(\Theta_{\pixel})} ,
\end{equation}
where $\normal(\pixel)$ is the normal vector of the 3D local surface computed by the deprojected points $\vertex(\pixel)$, $\Theta_{\pixel}$ is the set of all 2D neighboring pixels around $\pixel$, $G(\Theta_{\pixel})$ is an operator that counts the number of valid depth pixels (non-zero) in the neighborhood $\Theta_{\pixel}$, and at last, $\langle \cdot , \cdot \rangle$ is the inner product between 2 vectors. Note that $\normal(\pixel) $ is normalized so that  $|\langle \normal , \normal' \rangle| \in [0,1]$ and $|\cdot|$ is the absolute value operator.

\section{Experimental Results}
In this section we quantitatively and qualitatively demonstrate the effectiveness of our self-supervised approach against recent state-of-the-art supervised methods, as well as traditional filtering approaches.
The recently released Intel RealSense D415, an active stereo RGB-D sensor, is used for data collection and evaluation.

\begin{figure}[t]
    \centering
    \includegraphics[width=0.95\columnwidth]{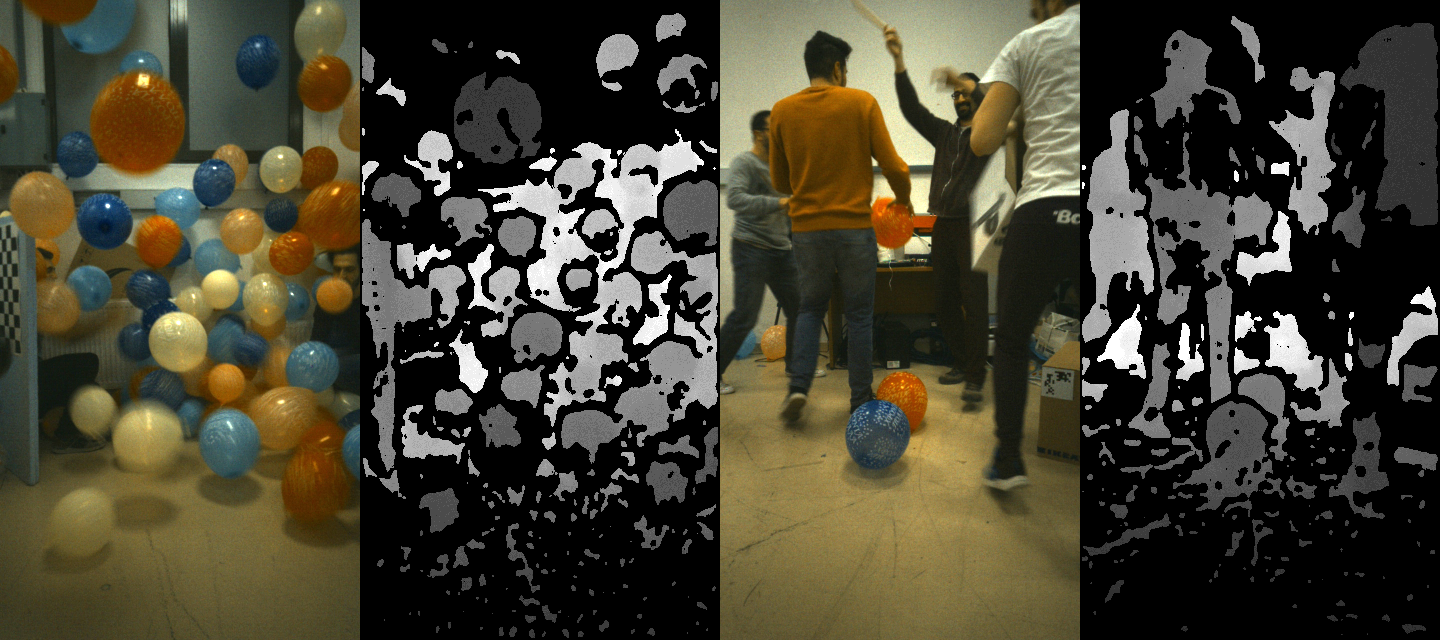}
    \caption{Collected training set samples showing the captured content (balloons and the multi-person activities).}
    \label{fig:training_set}
    \vspace{-0.2in}
\end{figure}

\textbf{Training RGB-D Dataset.} 
For training our model, a new RGB-D corpus has been collected, using multiple D415 devices containing more than 10K quadruples of RGB-D frames.
We employ $V=4$ vertically orientated sensors in a semi-structured deployment as depicted in Fig.~\ref{fig:intro_concept}, using a custom-made H-structure. 
The H-structure offers approximate symmetric placement and different vertical and horizontal baselines.
For the sake of spatio-temporal alignment between the color and the depth streams of the sensor, the infrared RGB stream was used instead of the extra RGB only camera.
This ensures the alignment of the color and depth image domains, and circumvents a technical limitation of the sensors that does not offer precise HW synchronization between the stereo pair and the RGB camera.
The sensor is configured to its ``high accuracy" profile, offering only high confidence depth estimates, but at the same time, producing highly sparse depth data, \textit{i.e.}~ $\approx60\%$ of zero-depth values in typical human capturing scenarios.
Data were captured using \cite{sterzentsenko2018lowcost},
spatial alignment between the 4 sensors was achieved by the multi-sensor calibration of \cite{papachristou2018markerless}, while, precise temporal alignment was achieved through the inter-sensor HW synchronization offered by the D415 sensors.

As our approach relies on view synthesis for supervision, we can easily collect raw data for training.
This is advantageous compared to using 3D reconstruction methods to generate near ground truth datasets \cite{Yan2018, Jeon2018}.
With respect to the dataset content, aiming to create a dataset of sufficient depth variability, we captured human actions simultaneously performed by multiple people as well as a special set of moving textured balloons of different colors. 
In detail, multiple subjects (1-7) performed free (\textit{i.e.}~not predefined) actions, while a variety of balloons were blown in the air using a blowing machine, creating depth maps of high variability. Note that the random movement patterns fully covered the sensors' FoV and prevented spatial bias in the training set. 
Indicative samples are depicted in Fig.~\ref{fig:training_set}.

\begin{table*}[ht]
\caption{Quantitative evaluation of the denoising algorithms: Depth-map and surface errors as well as errors in a 3D reconstruction task.}
\small
\vspace{0.05in}
\centering
\begin{tabular}{ l c c c c c c | c}
	\hline
	\multicolumn{4}{c}{\textbf{Euclidean Distance}} & \multicolumn{3}{c}{\textbf{Normal Angle Difference}} & \textbf{Kinect Fusion}\\
	\hline
	 & \textbf{MAE (mm)} & \textbf{RMSE (mm)} & \textbf{Mean (\degree) $\downarrow$} & \textbf{10.0 (\%) $\uparrow$} & \textbf{20.0 (\%) $\uparrow$} & \textbf{30.0 (\%) $\uparrow$} & \textbf{RMSE (mm)}\\
	\hline
	DDRNet \cite{Yan2018} & 114.57 & 239.06 & 52.85 & 1.78 & 7.30 & 16.59 & 50.79 \\
	DRR \cite{Jeon2018} & 75.40 & 201.49 & \textbf{30.23} & \textbf{10.95} & \textbf{34.69} & \textbf{57.76}   & 37.31\\
	JBF \cite{KCLU07} & 27.10 & 84.84 & 38.57 & 6.14 & 21.08 & 39.61 & 27.68 \\
	RGF \cite{Zhang2014} & 26.60 & 81.35 & 31.84 & 9.46 & 31.00 & 53.58 & 32.58 \\
	BF \cite{Tomasi} & 26.11 & 73.25 & 35.04 & 7.42 & 25.38 & 46.11 & 29.85 \\
	\hline
	\textbf{Ours} & \textbf{25.11} & \textbf{58.95} & 32.09 & 9.61 & 31.34 & 53.65 & \textbf{24.74} \\      
\end{tabular}
\vspace{-0.2in}
\end{table*}

\textbf{Implementation Details.} The training methodology along with the network hyper-parameters are presented in Section 1.1 of the supplementary material.

\begin{figure}[t]
    \centering
    \includegraphics[width=0.85\columnwidth]{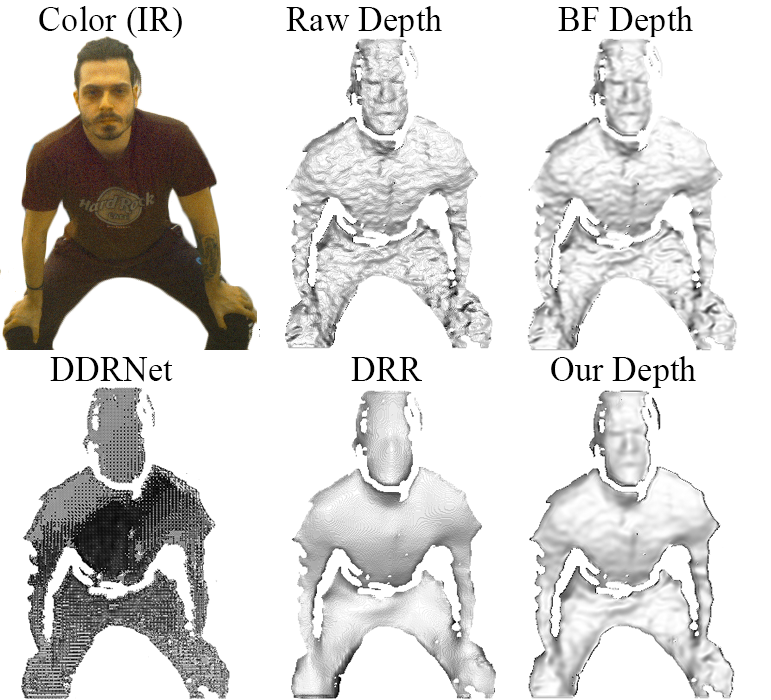}
    \caption{Qualitative results using D415 data.}
    \label{fig::ddepthmap_result}
    \vspace{-0.2in}
\end{figure}

\textbf{Evaluation Methodology}.
The proposed model is evaluated against traditional filtering methods, such as Bilateral Filter (BF~\cite{Tomasi}), Joint Bilateral Filter (JBF~\cite{KCLU07}), Rolling Guidance (RGF~\cite{Zhang2014}), as well as data-driven approaches such as DRR~\cite{Jeon2018} and DDRNet~\cite{Yan2018}. 
Note that for the DDRNet case, the refinement part of the network is omitted in order to have a fair comparison in denoising. 
Due to the lack of ground truth, depth maps from Kinect v2 (K2) \cite{sell2014xbox} are used as ``close to ground truth" data for the quantitative evaluation. 
That is, a 70-pair RGB-D set of instant samples with varying content are captured using a rigid-structure that combines K2 and D415, and is used as test set for evaluation purposes. 
In particular, to achieve the closest possible positioning between the modalities, the two sensors are placed in a way that the overlap of their FoV is high, while the structure is calibrated using the Matlab Stereo Camera Calibrator App \cite{zhang2000flexible}. 
The evaluation consists of 3 experiments varying from direct depth map comparison to application-specific measurements. 
In detail, for the first experiment the depth maps captured by the D415 sensor are denoised by the proposed network and the state-of-the-art methods and the result is evaluated using the K2 ground truth data. 
Subsequently, using the aforementioned rigid-structure, we capture 15 scanning sequences with both sensors (D415, K2) simultaneously, which are then utilized as inputs to KinectFusion. 
For our last experiment, we utilize a multi-view setup to capture 5 full-body samples.
Note that besides quantitative evaluation, for each experiment qualitative results are also presented.

\begin{figure}[t]
    \centering
    \includegraphics[width=0.85\columnwidth]{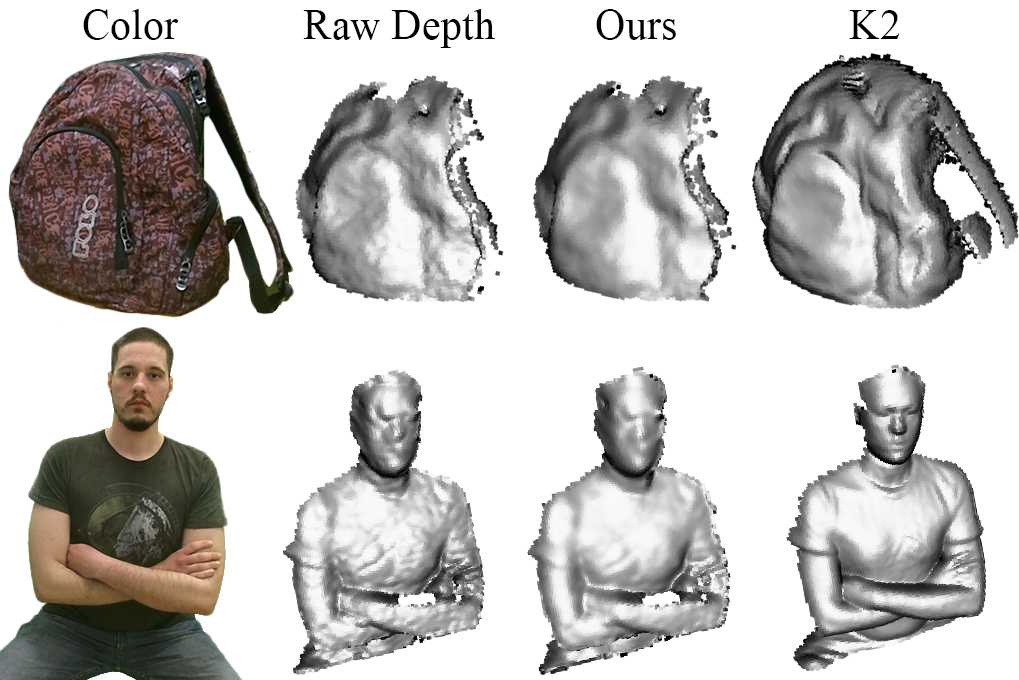}
    \caption{Qualitative results using KinectFusion.} 
    \label{fig::KinectFusion}
    \vspace{-0.2in}
\end{figure}

\textbf{Results.} 
In the first experiment we use projective data association to compare the performance of denoising methods on D415 data against the close to ground truth K2 depth maps.
The results are presented in Table \ref{table:comparison} (columns 2-7) and showcase the effectiveness of the proposed method against supervised methods and traditional filters.
Despite the low mean absolute error differences between the quantified methods, the RMSE results prove the effectiveness of our approach to denoise depth maps by achieving the lowest error deviation. 
Regarding surface errors, our method ranks third following DRR \cite{Jeon2018} and RGF \cite{Zhang2014} with slight differences. 
However, DRR filtering results in depth map oversmoothing and spatial offsets (bendings), degenerating high frequent details, thus causing large distance errors. 
On the other hand, DDRNet \cite{Yan2018} under-performs in D415 depth data denoising. 
This can be attributed either to the context specific learning of the network on high density depth maps of humans without background, which showcases the disadvantage of using specific 3D reconstruction methods to generate near ground truth data for supervision, and the inability to generalize well. 
Another reason may be that the noise level of D415 is higher than the sensors \cite{Yan2018} was trained with.
In addition, the fact that D415 produces sparse results hampers the applicability of CNN-based methods that did not account for that, due the nature of convolutional regression. 
Finally, classical and guided filters present comparatively larger errors than the proposed method. 
Qualitative results\footnote{Additional qualitative results related to our experiments are included in the supplementary material document.} of the depth map denoising are illustrated in Fig.~\ref{fig::ddepthmap_result}.
It is apparent that local filtering cannot sufficiently denoise due to its local nature, while the learning-based alternatives either oversmooth (DRR) or fail to generalize to other sensors (DDRNet). Instead, our approach smooths out the noise while preserving structural details.

\begin{figure}[t]
    \centering
    \includegraphics[width=0.9\columnwidth]{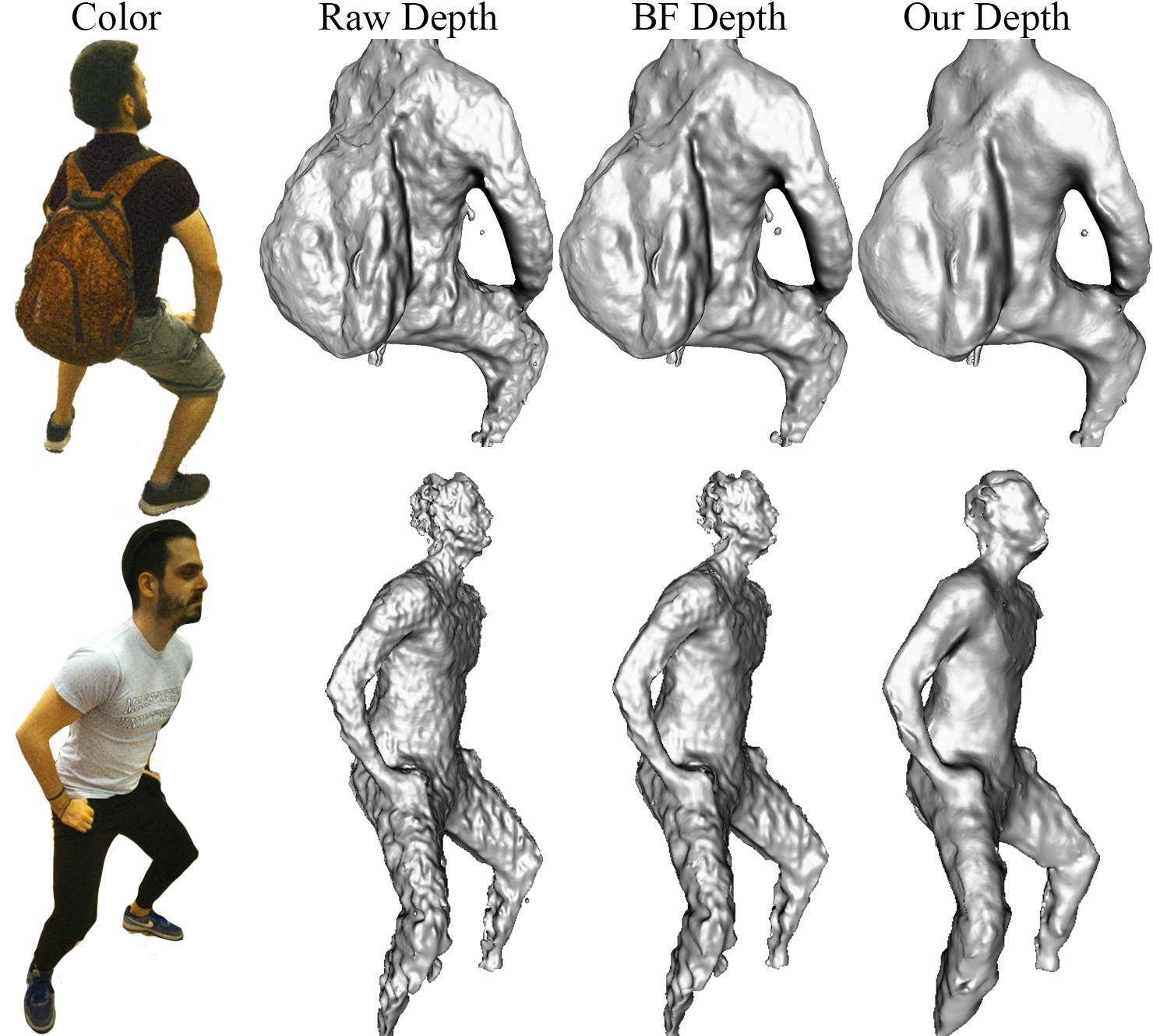}
    \caption{Qualitative results using Poisson reconstruction.} 
    \label{fig::3d_reco}
    \vspace{-0.1in}
\end{figure}

\begin{figure}[t]
    \centering
    \includegraphics[width=0.95\columnwidth]{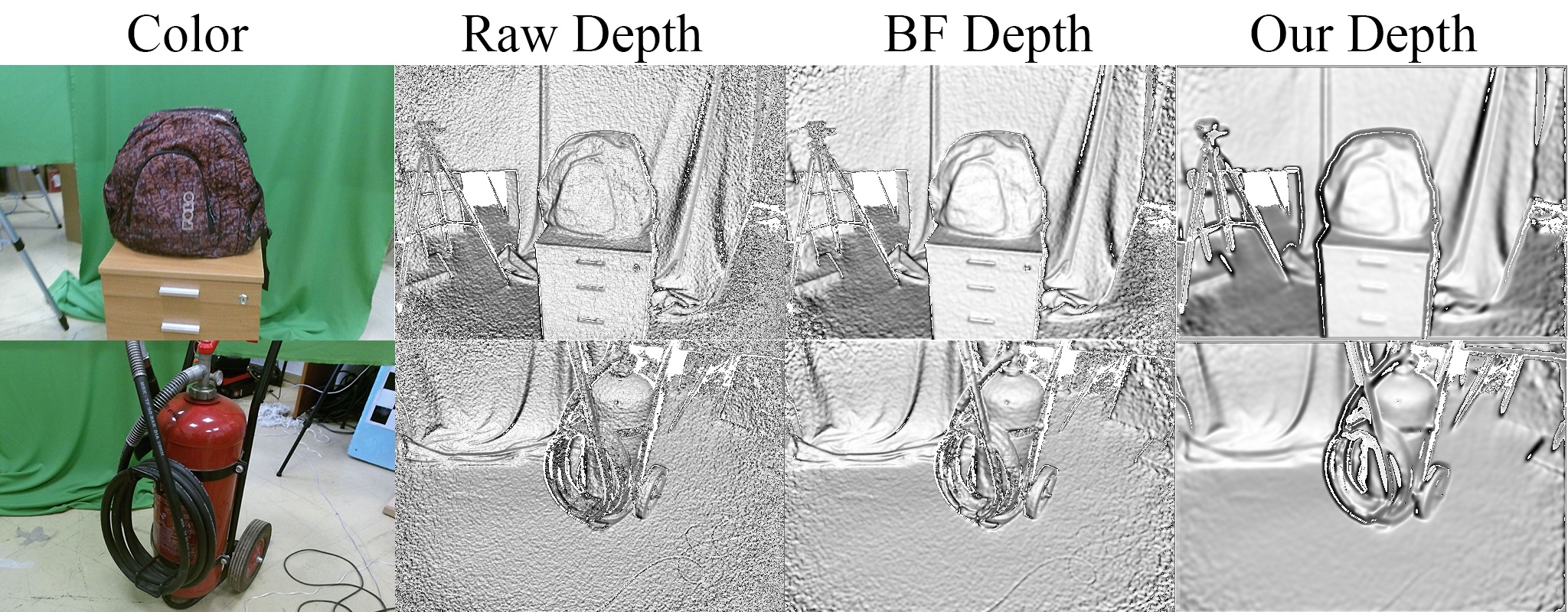}
    \caption{Qualitative comparison using K2 data.}
    \label{fig:k2_results}
    \vspace{-0.2in}
\end{figure}

The second experiment demonstrates results in an application setting using KinectFusion to 3D reconstruct static scenes.
The rationale behind this experiment is the comparison of the scanning results using the denoised depth maps of D415 and comparing the result with that of a K2 scan.
Quantitative results are presented in Table~\ref{table:comparison} (last column), while qualitative results are illustrated in Fig.~\ref{fig::KinectFusion}.
In this experiment we opt to use an aggregated metric that handles surface and geometry information jointly, point-to-plane.
Instead of relying on the nearest neighbor for distance computation, we calculate the Least Square Planes for each point in the close-to-ground truth point cloud using all vertices in a 5mm radius (a 2mm voxel grid size was used when 3D scanning). The distance of each calculated plane of the ground-truth point cloud against the closest point from the denoised point clouds contribute a term to the final RMSE.

While KinectFusion reconstructs surfaces by aggregating and fusing depth measurements it also implicitly denoises the result through the TSDF fusion process. 
In order to accentuate surface errors we conduct another experiment, this time using Poisson reconstruction \cite{kazhdan2013screened}, which requires better surface estimates in order to appropriately perform 3D reconstruction. 
This allows us to qualitatively assess the denoised output smoothness, while also showcasing the preservation of structure.
We spatially align 4 D415 sensors in a $360^\circ$ placement and capture depth frame quadruples of static humans.
We use deprojected raw and denoised depth maps to point clouds and calculate per point normals using the 10 closest neighbors. These oriented point clouds are reconstructed using \cite{kazhdan2013screened} with the results illustrated in Fig.~\ref{fig::3d_reco}.
It is apparent that BF, one of the performing filters of the first experiment, performs smoothing without removing all noise as it operates on local level. 
On the contrary, the 3D reconstructed model using the denoised depth maps of our model achieves higher quality results, mainly attributed to its ability to capture global context more effectively.

Finally, while other denoising CNNs trained using other sensors fail to produce good results on D415, we also present qualitative results on K2 data\footnote{We collect our own data as the K2 dataset of \cite{Yan2018} is not yet publicly available.}, albeit trained using D415 noisy depths.
Fig.~\ref{fig:k2_results} shows that our model gracefully handles noise from other sensors, contrary to fully supervised methods that are trained on datasets of a specific context (sensor, content).

\section{Conclusion}
In this paper, an end-to-end model was presented for the depth denoising task. To tackle the lack of ground truth depth data, the model was trained using multiple RGB-D views of the same scene using photometric, geometrical, and surface constraints in a self-supervised manner. The model outperformed both traditional filtering and data-driven methods, through direct depth map denoising evaluation and two well-established 3D reconstruction applications. Further, it was experimentally shown that our model, unlike other data-driven methods, maintains its performance when denoising depth maps captured from other sensors. The limitations of the method lie in the need of color information for supervision, and sensors' hardware synchronization.\\

\noindent \textbf{Acknowledgements.} We thank Antonis Karakottas for his help with the hardware setup, and the subjects recorded in our dataset for their participation. We also acknowledge financial support by the H2020 EC project VRTogether under contract 762111, as well as NVidia for GPU donation.

{\small
\bibliographystyle{ieee_fullname}
\balance
\bibliography{egbib}
}
\appendix
\newpage
\phantomsection
\addcontentsline{toc}{chapter}{Appendices}
\begin{huge}
\textbf{Supplementary}
\end{huge}
\setcounter{section}{0}
\renewcommand{\thesection}{\Alph{section}}
\renewcommand{\theHsection}{appendixsection.\Alph{section}}
\nobalance
\section{Introduction}
\begin{figure}[t]
\begin{center}
   \includegraphics[width=0.8\linewidth]{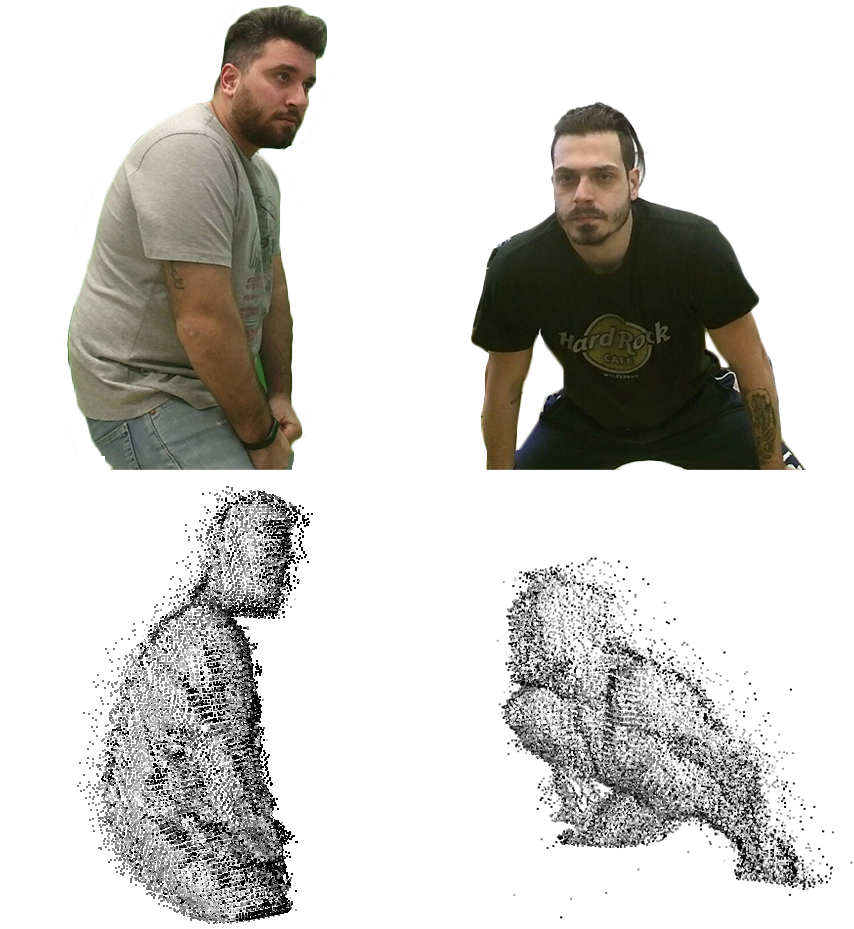}
\end{center}
   \caption{Denoising results using original DDRNet\cite{Yan2018} model on Kinect 2 data.
   }
\label{fig:ddrnet_kinect}
\end{figure}

In this supplementary material we complement our original manuscript with additional quantitative and qualitative results, which better showcase the advantages of the proposed self-supervised denoising model over traditional filtering and supervised CNN-based approaches.
In particular, we present the adopted implementation details used for training our model, as well as additional qualitative results for the two 3D application experiments presented in the original manuscript, namely 3D scanning with KinectFusion \cite{newcombe2011kinectfusion} and full-body 3D reconstructions using Poisson 3D surface reconstruction \cite{kazhdan2013screened}. A comparative evaluation with the learning-based state-of-the-art methods on InteriorNet (IN)~\cite{InteriorNet18} follows, while an ablation study concludes the document.\\
\indent The aforementioned results based on all methods presented in the originally manuscript, namely Bilateral Filter (BF~\cite{Tomasi}), Joint Bilateral Filter (JBF~\cite{KCLU07}), Rolling Guidance (RGF~\cite{Zhang2014}), and data-driven approaches (DRR~\cite{Jeon2018}, DDRNet~\cite{Yan2018}).
Note that for the DRR and DDRNet methods, additional results aim to highlight the over-smoothing effect of the former and the weakness of the latter to denoise depth maps captured by the Intel RealSense D415 sensors.
In more detail, DRR is trained on static scenes that contain dominant planar surfaces and, thus tends to flatten (\textit{i.e.}~over-smooth) the input data.
On the other hand, the available DDRNet model \footnote{\url{https://github.com/neycyanshi/DDRNet}} that we used, produces high levels of flying pixels (\textit{i.e.}~spraying, see Fig.~\ref{fig:ddrnet_kinect}) which can be attributed to background (zero depth values) and foreground blending, even though its predictions are appropriately masked.
While the authors have not provided the necessary information, it is our speculation that the available model is trained using Kinect 1 data, which is partly supported by the sub-optimal results it produces on Kinect 2 data. 

Qualitatively, the remaining traditional filters (BF, JBF, RGF) perform similarly, with RGF showcasing the most competitive results to our method. However, note that RGF utilizes color information in an iterative scheme. It is worth mentioning that depending on the evaluation, \textit{i.e.}~KinectFusion or Poisson reconstruction, the difference in the quality among the methods may be more or less distinguishable. \\

\subsection{Implementation Details}
The CNN-based autoencoder presented in the original manuscript is implemented using the PyTorch framework \cite{pytorch}. The hyperparameters' initialization follows, while the notation from Section 3.2 (see original manuscript) is adopted.
We set $\lambda_{1}=0.85, \lambda_{2}=0.1, \lambda_{3}=0.05$, while $\alpha$ for the photometric loss is set to $0.85$. 
Regarding the outlier estimators, we set $\gamma=0.447$ for the Charbonnier and $c=2.2$ for the Tukey penalty, respectively. 
During training, ELU(a) non-linearity with $a=1$ is used for all CONV layers except for the output one, while Adam \cite{adam} with $\beta_{1}=0.9, \beta_{2}=0.99$ is used for optimization. 
Xavier initialization \cite{glorot2010understanding} is used for the network weights.
The network is trained with learning rate set to $0.0002$ and a mini-batch size of 2. 
Training converges after about $102k$ iterations. 
The network is trained with depth and color images of $640\times 360$ resolution, while no data augmentation is performed. 
All collected depth maps are thresholded to $3m$ and thus $\sigma_{D} = 3$.
Note that the mean inference time on a GeForce GTX 1080 graphics card is 11ms. 

\subsection{KinectFusion Reconstruction}
\label{sec:kinfu}
KinectFusion \cite{newcombe2011kinectfusion} reconstructs 3D surfaces by temporally aggregating  and  fusing  depth  maps, also  implicitly denoising the outcome through the Truncated Signed Distance Functions (TSDFs) fusion process. 
Therefore, the method's most typical failure case corresponds to an insufficient registration of an input frame, either attributed to difficult to track motion (pure rotation, fast translation) or noisy input.
Our results are offered in the exact same sequences, and thus the former source of error is removed, with any tracking failures attributed to noisy inputs.

\begin{figure*}[!ht]
    \centering
    \includegraphics[width=.65\textwidth]{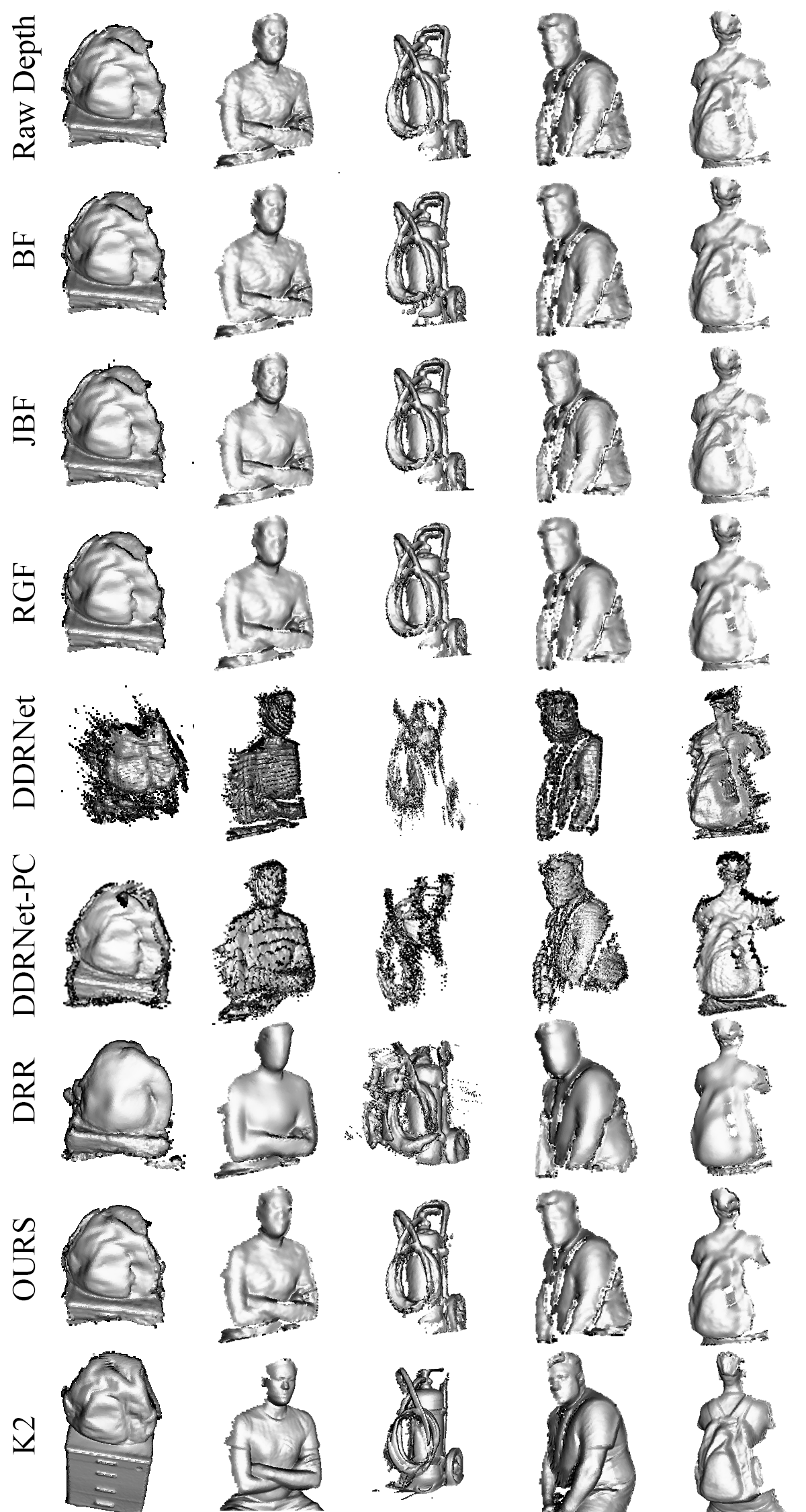}
    \caption{Reconstruction results of KinectFusion scans. It is worth mentioning that even noisy raw input can be reconstructed into a high quality mesh (row 1). DDRNet and DRR fail to produce adequate quality meshes (see Section \ref{sec:kinfu}). Our model along with K2 and RGF produce the best qualitative results, preserving a fair amount of structural details (\textit{e.g.}~face, bag, folds).}
    \label{fig:kinfu}
\end{figure*}

Consequently, even noisy depth observations may result in high quality 3D scans. 
Although the original depth estimates from D415 are noisy, in most cases, KinectFusion manages to reconstruct a relatively smooth 3D mesh surface. 
This is illustrated in Fig. \ref{fig:kinfu} in the first row, where the resulting meshes using the raw depth input are presented. 
It is worth noting that D415 depth map denoising is proven challenging for the data-driven methods. In particular, DRR tends to over-smooth the surfaces, while DDRNet is completely incompatible with the depth data.

KinectFusion on DDRNet denoised data was repeatedly failing to make correspondences in consecutive frames due to the increased amount of ``spraying'' in the denoised output. Thus, the fact that our proposed method does not fall into the same limitations as the other data-driven methods can be considered an advantage.
KinectFusion-based results are shown in Fig.~\ref{fig:kinfu}. For comparison, the last row of Fig.~\ref{fig:kinfu} shows 3D reconstruction using frames acquired by Microsoft Kinect 2 device, which captures higher quality depth.

We further experimented with DDRNet by re-implementing its denoising part, using traditional and partial convolutions, denoted as DDRNet-TC and DDRNet-PC, respectively. The model was trained using our dataset, which resulted in better results due to the sparse nature of the data. Since our dataset does not contain ground-truth depth-maps, we employ forward-splatting (see Section 3.1 of the original original manuscript) in order to produce cleaner depth-maps to use as near ground-truth. As Table~\ref{tbl:res}(top) shows, our model outperforms DDRNet retrained models, in both regular and partial convolution by a wide margin, which can be attributed to the different behaviour of splatting color images compared to depth maps. For completeness, we qualitatively evaluated the performance of DDRNet-PC using KinectFusion (see Fig.~\ref{fig:kinfu}, middle). Note that even if denoising is improved compared to the original DDRNet, the reconstructed output quality is still low.

\begin{figure}[t]
\begin{center}
   \includegraphics[width=0.8\linewidth]{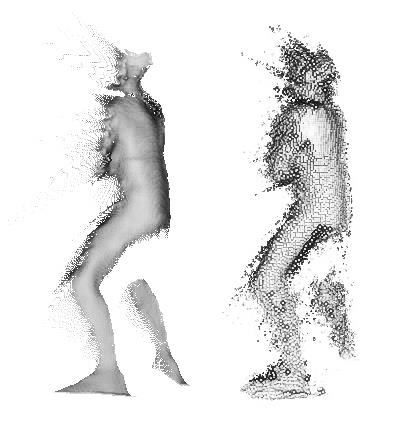}
\end{center}
\vspace{-0.2in}
   \caption{Projected depth maps to 3D domain, after denoising with DRR \cite{Jeon2018} (left) and original DDRNet \cite{Yan2018} (right). This figure showcases the spray at the boundaries, which leads Poisson reconstruction method to fail (see Section
   \ref{sec:poisson}).
   }
\label{fig:pointcloud1}
\end{figure}

\subsection{Poisson Reconstruction}
\label{sec:poisson}
\begin{figure*}[!ht]
    \centering
    \includegraphics[width=.9\textwidth]{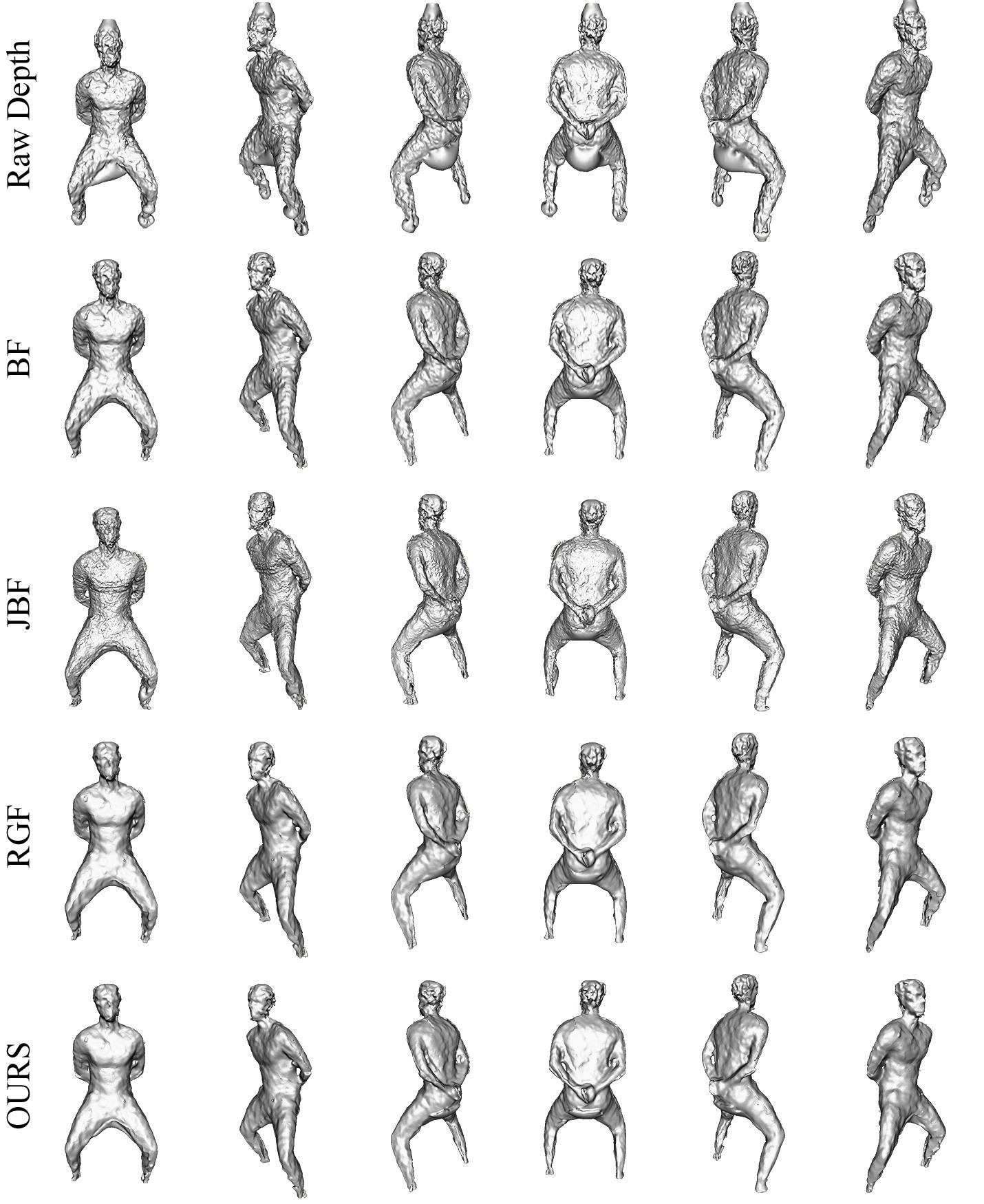}
    \caption{Poisson reconstruction sample. BF and JBF lead to low quality reconstruction due their inability to understand the global context of the scene. Our method and RGF lead to higher quality reconstructions, restoring face details that are hardly spotted in ``Raw Depth" reconstruction.}
    \label{fig:Poisson1}
\end{figure*}

\begin{figure*}[!ht]
    \centering
    \includegraphics[width=.9\textwidth]{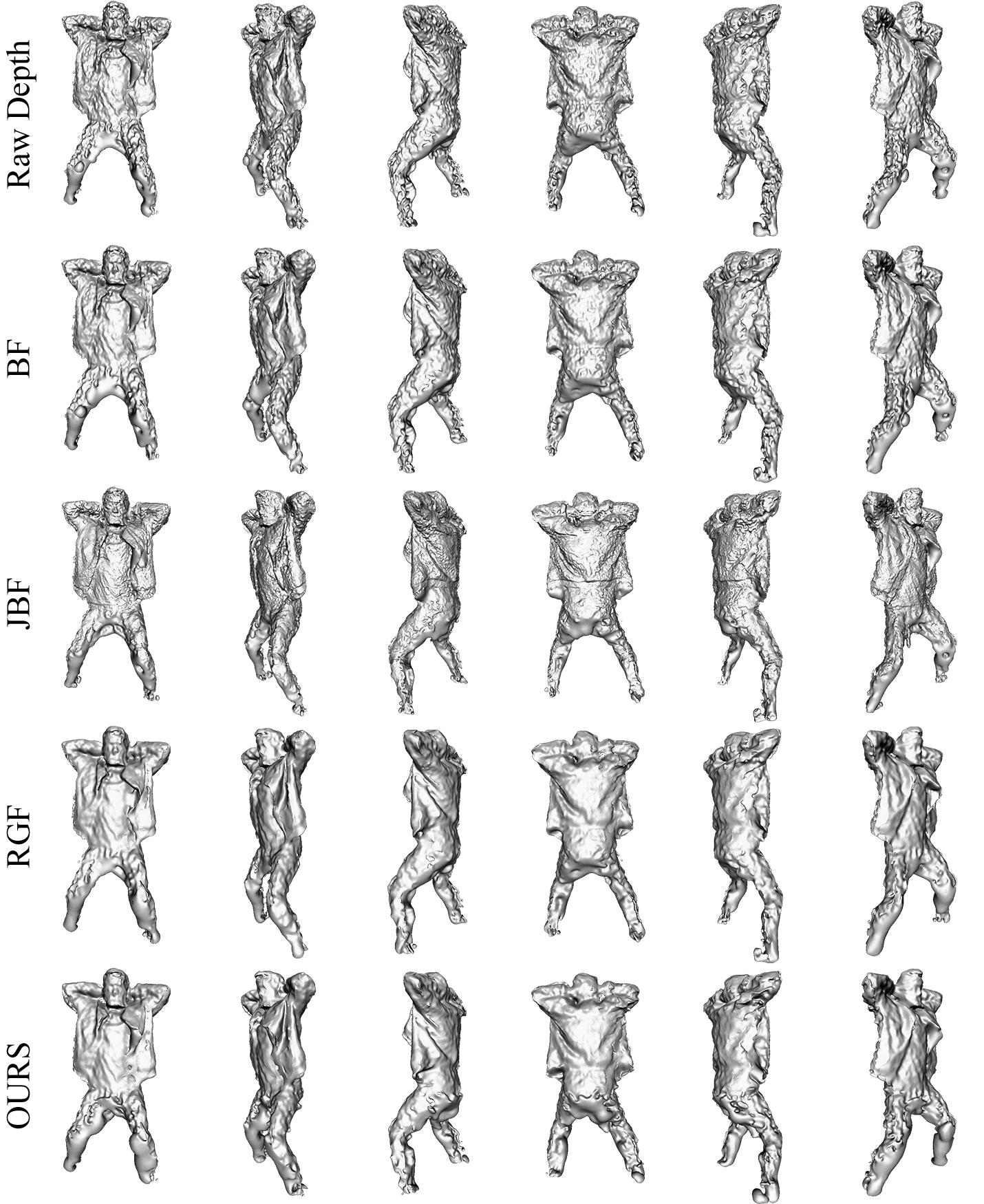}
    \caption{Poisson reconstruction sample with the sensors placed higher (looking downwards) and slightly further away from the target. This leads to erroneous surface estimations at the approximate leg region, mainly due to the partial visibility and data sparseness. Despite the challenging setup, our method was able to successfully remove noise and preserve fine details (\textit{e.g.}~face, jacket folds). }
    \label{fig:Poisson2}
\end{figure*}

\begin{figure*}[!ht]
    \centering
    \includegraphics[width=.9\textwidth]{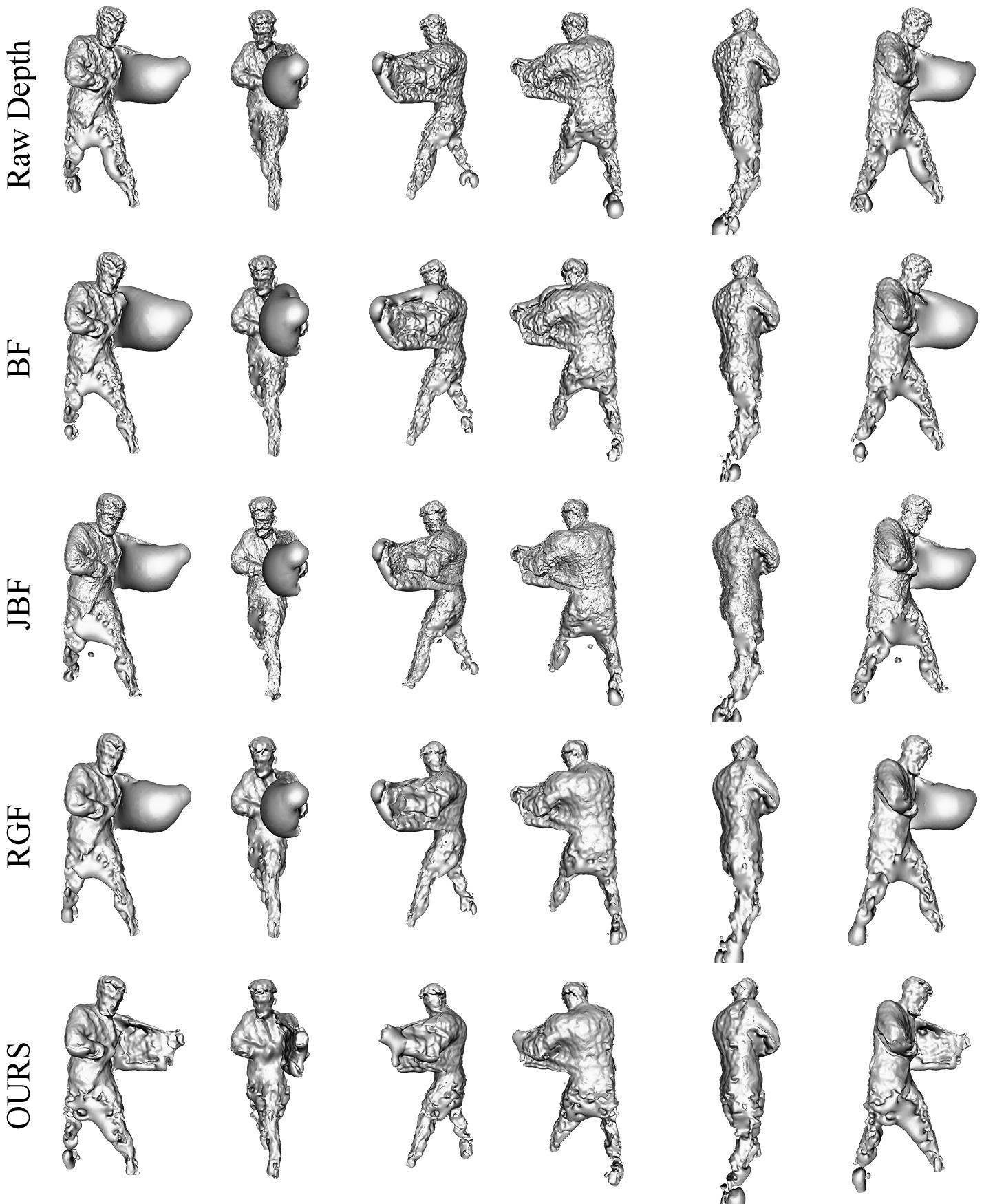}
    \caption{Poisson reconstruction sample using the setup described in  Fig.~\ref{fig:Poisson2}. Our method is the only one to remove the ``balloon" noise at the inner side of the jacket caused by noisy depth measurements.}
    \label{fig:Poisson3}
\end{figure*}

The second method used to qualitatively compare the aforementioned methods is the well-established 3D Poisson reconstruction \cite{kazhdan2013screened}.
The setup is realized as 4 RealSense D415 sensors placed in a cross-like setup to capture a static subject in a full $360^\circ$ manner. 
Poisson reconstruction utilizes surface information (oriented point-clouds) in order to recover the original 3D shape, constituting an appropriate application to compare denoising results while preserving geometric details in a qualitative manner.
While the produced reconstructions are watertight, ``balloon" like artifacts can be observed in empty areas where the proximal surface information is inconsistent or noisy.
This can be seen in Fig.~\ref{fig:Poisson1} (1st row-``Raw Depth").
The curvature of these proximal patches is an indicator of the smoothness across the boundary of the hole (empty area).
Fig.~\ref{fig:Poisson1}-\ref{fig:Poisson3} demonstrate the results of full-body 3D reconstructions using $4$ depth maps denoised with each evaluated method, as well as the original raw input (1st row).
Note that results using DDRNet (both original and retrained) and DRR methods are omitted, as their denoised depth maps are affected by ``boundary spraying" (see Fig.~\ref{fig:pointcloud1}), which leads to highly cluttered 3D reconstructions.
Depth maps denoised using RGF are also affected by slight spraying, which is easily removed manually in order to present a fair quality result. 
From the presented results, it can be seen that 3D reconstruction from raw (noisy) depth maps preserves little to no geometric details, while using depth maps from BF and JBF methods leads to local region smoothing. 
On the other hand, RGF leads to higher quality results as which are comparable to those produced by our model.
It should be noted though that our model infers using only depth input while RGF requires color information and an appropriate selection of parameters.

\begin{table}[t]
\caption{Top to bottom sections: a) DDRNet trained with splatted depth, b) learning-based methods evaluation on IN, c) ablation results of the proposed denoising model.}
\label{tbl:res}
\small
\vspace{-0.1in}
\setlength\tabcolsep{1.5pt} 
\renewcommand{\arraystretch}{0.85}
\begin{center}
\begin{tabular}{l c c c c c c}
\hline
\\[-0.75em]
\textbf{Model} & \textbf{MAE} & \textbf{RMSE} & \textbf{M(\degree)$\downarrow$} & \textbf{10(\%)$\uparrow$} & \textbf{20(\%)$\uparrow$} & \textbf{30(\%)$\uparrow$}\\
\\[-0.75em]
\hline
\\[-0.75em]
DDRNet-TC & 121.83 & 265.10 & 56.17 & 1.41 & 5.79 & 13.41\\
DDRNet-PC & 75.68 & 241.58 & 40.46 & 5.38 & 18.83 & 36.13\\
\hline

\\[-0.75em]
DDRNet (IN) & 140.80 & 198.45 & 59.86 & 1.72 & 6.07 & 11.32\\
DRR (IN) & 86.88 & 144.97 & 25.84 & 26.72
 & 48.62 & 65.19 \\
\textbf{Ours (IN)} & \textbf{33.44} & \textbf{81.28} & \textbf{20.08} & \textbf{39.53} & \textbf{64.12} & \textbf{77.37}\\
\hline
\\[-0.75em]
AE & 26.35 & 59.92 & 36.30 & 7.78 & 25.75 & 45.70\\
P+N & 28.04 & 60.20 & 34.32 & 8.73 & 28.55 & 49.46\\
P+D & 26.43 & \textbf{58.31} & \textbf{31.71} & \textbf{9.62} & \textbf{31.39} & \textbf{53.98} \\
P-only & 25.96 & 58.30 & 32.13 & 9.39 & 30.69 & 53.11\\
P+D+N (best) & \textbf{25.11} & 58.95 & 32.09 & 9.61 & 31.34 & 53.65\\
\\[-0.75em]
\hline
\end{tabular}
\end{center}
\vspace{-0.2in}
\end{table}

\subsection{Evaluation on InteriorNet}
\begin{figure*}[!ht]
    \centering
    \includegraphics[width=\textwidth]{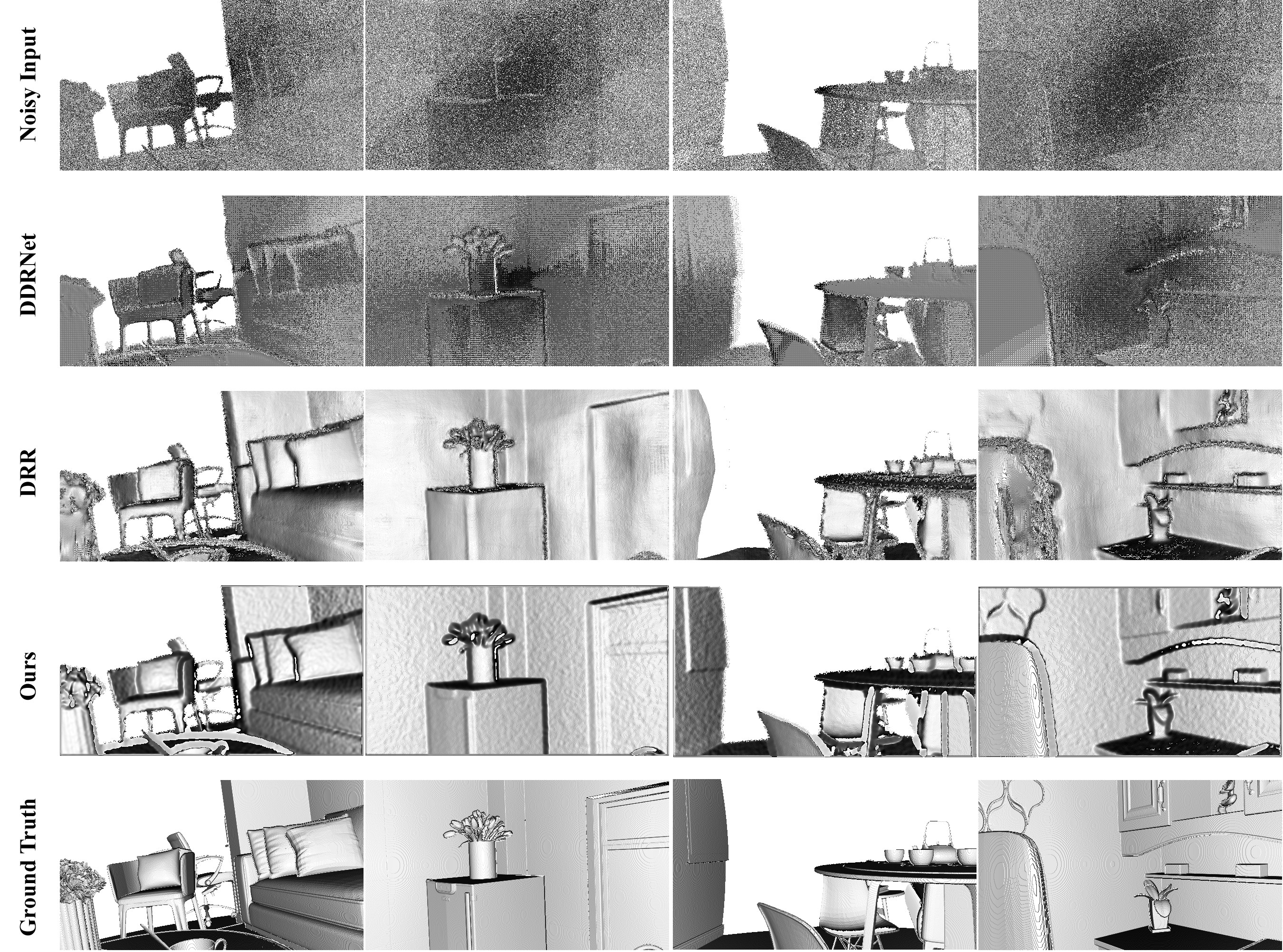}
    \vspace{0.05in}
    \caption{Quantitative results of learning-based methods in rendered images from InteriorNet layouts. The first row and last row show the noisy input and ground truth respectively. DDRNet fails to remove most of the noise. On the other hand, DRR shows promising results, although it tampers the shape of some objects and fails to preserve fine details. Our model shows superior results comparing to other methods on these data.}
    \label{fig:InteriorNet}
\end{figure*}

IN consists of 22M layouts of synthetic indoor scenes with varying lighting configuration. We use 6K samples from the first 300 scenes. We corrupt these clean ground truth depth maps with two artificial noise patterns in order to create noisy-ground truth data pairs; a) a noise similar to the one presented in~\cite{barron2013intrinsic}, and b) a ToF-like, non-linear (distance-dependent), bi-directional noise distribution along the ray. The quantitative results of the learning - based methods are shown in Table~\ref{tbl:res}(middle). As for qualitative results on this task, we provide the original, ground truth and denoised images in Fig.~\ref{fig:InteriorNet}.

\subsection{Ablation Study}
Finally, we perform an ablation study of various aspects of out deep depth denoising model. Spacifically, we examine cases of a) training the model as a plain autoencoder (AE) without bell and whistles (only reconstruction loss used), b) training the AE with photometric loss only (P-only), c) regularizing supervision using the BerHu depth loss (P+D), d) using normal priors to guide the supervision (P+N), instead of depth regularization, and e) combining photometric supervision with depth and surface normals losses (P+D+N), as presented in the original manuscript. 

The results of the aforementioned cases evaluated on our dataset are presented in Table~\ref{tbl:res}(bottom). These results indicate that photometric supervision is a better alternative than a plain autoencoder train with a reconstruction loss, as well as that depth regularization is important as it aids photometric supervision by constraining it when its assumptions break (no texture, etc.). Further, note that the normals smoothness prior leads to a significant improvement of the MAE, while achieving the second best performance in the rest of error metrics. Based on this analysis, we adopt the last training scheme for our depth denoising model.

\end{document}